\documentclass[10pt,twocolumn,letterpaper]{article}

\usepackage[pagenumbers]{cvpr} % To force page numbers, e.g. for an arXiv version

\usepackage{booktabs}

\usepackage{times}
\usepackage{epsfig}
\usepackage{graphicx}
\usepackage{amsmath}
\usepackage{amssymb}
\usepackage{bbm}
\usepackage{pifont}
\usepackage{subcaption}
\usepackage{mathtools}
\usepackage{multirow}
\usepackage{adjustbox}
\usepackage{dsfont}
\usepackage[utf8]{inputenc}
\usepackage{comment}
\usepackage{color, colortbl}
\usepackage[accsupp]{axessibility}

\usepackage[dvipsnames]{xcolor}
\usepackage{xspace}

\usepackage[pagebackref,breaklinks,colorlinks,citecolor=LimeGreen]{hyperref}

\usepackage[capitalize]{cleveref}
\crefname{section}{Sec.}{Secs.}
\Crefname{section}{Section}{Sections}
\Crefname{table}{Table}{Tables}
\crefname{table}{Tab.}{Tabs.}

\newcommand{\fabio}[1]{{\color{ForestGreen}[F: #1]}}

\newcommand{\real}{{\rm I\!R}}
\newcommand{\myparagraph}[1]{\vspace{4pt}\noindent\textbf{#1}}

\newcommand{\cmark}{\ding{51}}

\newcommand{\detector}{\mathcal{F}}
\newcommand{\data}{\mathcal{D}}

\def\cvprPaperID{6} % *** Enter the CVPR Paper ID here
\def\confName{CLVISION 2022}
\def\confYear{2022}

\begin{document}

\title{Modeling Missing Annotations for Incremental Learning in Object Detection}

\author{
Fabio Cermelli$^{1,2}$, Antonino Geraci$^{1}$, Dario Fontanel$^{1}$, Barbara Caputo$^{1}$\\
$^1$Politecnico di Torino, $^2$Italian Institute of Technology\\
{\tt\small fabio.cermelli@polito.it} \\
}

\maketitle

\begin{abstract}
Despite the recent advances in the field of object detection, common architectures are still ill-suited to incrementally detect new categories over time. They are vulnerable to catastrophic forgetting: they forget what has been already learned while updating their parameters in absence of the original training data.
Previous works extended standard classification methods in the object detection task, mainly adopting the knowledge distillation framework. However, we argue that object detection introduces an additional problem, which has been overlooked. While objects belonging to new classes are learned thanks to their annotations, if no supervision is provided for other objects that may still be present in the input, the model learns to associate them to background regions.
We propose to handle these missing annotations by revisiting the standard knowledge distillation framework. Our approach outperforms current state-of-the-art methods in every setting of the Pascal-VOC dataset. We further propose an extension to instance segmentation, outperforming the other baselines.
% The code will be released.
\end{abstract}

%%%%%%%%% BODY TEXT
\section{Introduction}
\label{sec:introduction}
% \begin{itemize}
%     \item OD is important
%     \item However, OD methods are not able to perform incrementally
%     \item ILOD
%     \item Especially, no one considered the background shift as in MiB
%     \item We propose UCE, UKD and a new attention on features.
%     \item Summary of Contributions (Sota on VOC, \textit{COCO ?}, InstanceSeg VOC)
% \end{itemize}

Object detection is a key task in computer vision that has seen significant development in recent years.
% The improvements have been achieved mainly thanks to the rise of deep neural network architectures \cite{girshick2015fast, ren2015faster, he2017mask, yolov3, lin2017focal, carion2020end} that improved results while reducing the computation.
% Despite the advances, these architectures assume to know all the classes that they will see a-priori and they are not suited to incrementally update their knowledge over time to learn new classes.
The advances were made possible by the rise of deep neural network architectures \cite{girshick2015fast, ren2015faster, he2017mask, yolov3, lin2017focal, carion2020end}, which improved results while reducing computation time.
Despite the advances, these architectures assume that they already know all of the classes they will encounter and are not designed to incrementally update their knowledge to learn new classes over time.
% A na\"ive solution would be restarting the training process from scratch, collecting a new dataset with all the classes and training again the architecture. However, this is impractical since it requires a huge computational overhead to re-learn the already known classes and it would require to use previous training data that may be no longer available, \eg due to privacy concerns or intellectual property. 
A na\"ive solution would be to restart the training process from the beginning, gathering a new dataset with all of the classes and retraining the architecture. However, this is impractical because it would necessarily require a significant computational overhead to re-learn the previously learned classes, as well as the use of previous training data that may no longer be available, for example due to privacy concerns or intellectual property rights.

\begin{figure}[t]
    \centering
    \includegraphics[width=\linewidth]{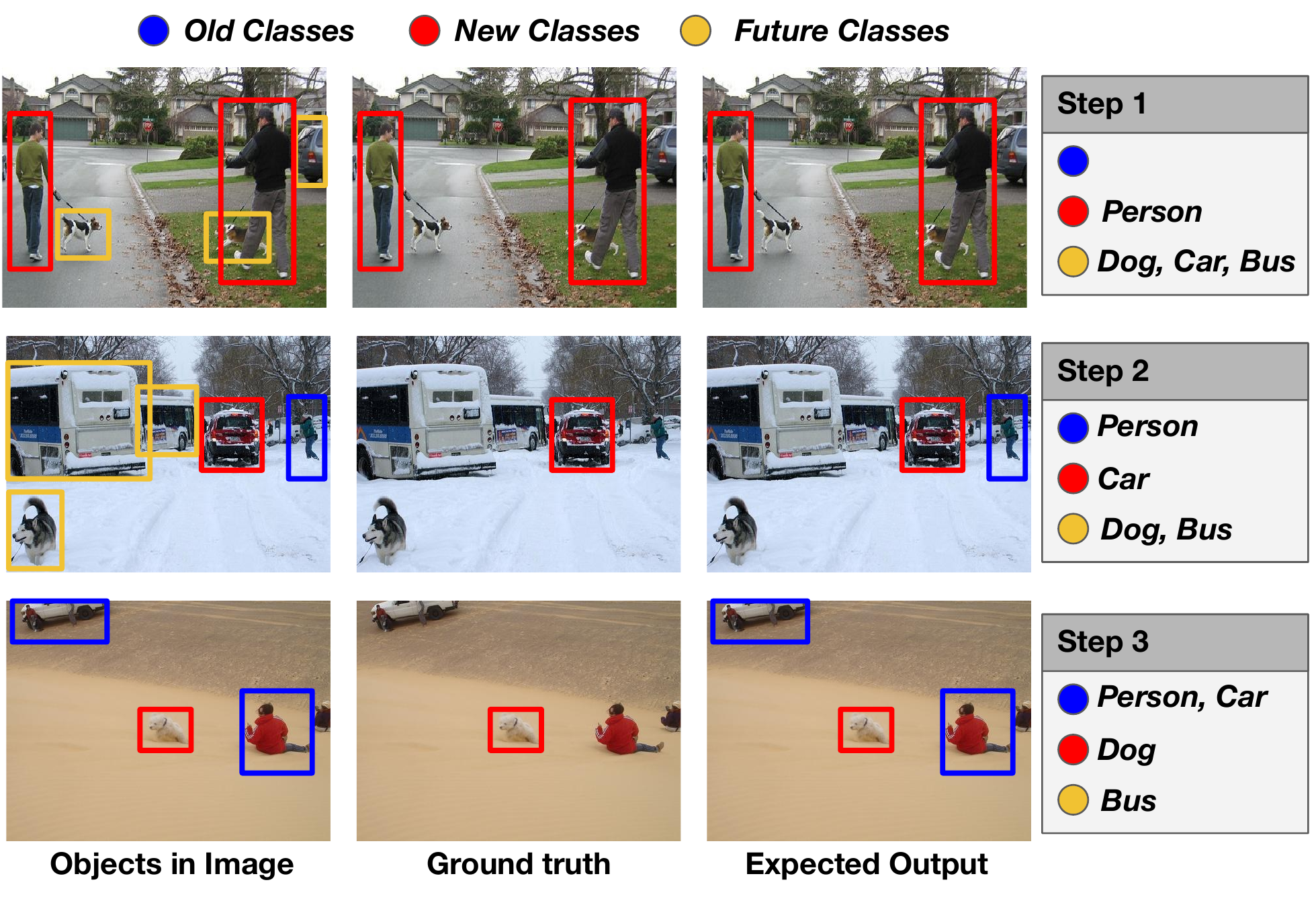}
    \caption{An illustration of the missing annotation issue of object detection in different time steps. At training step $t$, the annotations are provided only for new classes (red), while all the other objects, both from old (blue) and future (yellow) steps are not annotated.} \vspace{-8pt}
    \label{fig:teaser}
\end{figure}

% A more suited solution is to resort to incremental learning \cite{li2017learning} and directly update the models to extend their knowledge on the new classes by training only on new data while avoiding catastrophic forgetting \cite{mccloskey1989catastrophic}.
% Incremental learning has been largely studied in the context of image classification \cite{rebuffi2017icarl, douillard2020podnet, li2017learning, fini2020online, aljundi2018memory, rusu2016progressive, kirkpatrick2017overcoming}, but only recently it has been extended to more complex tasks such as object detection \cite{shmelkov2017incremental, peng2020faster, hao2019end, chen2019new, zhou2020lifelong} and semantic segmentation \cite{cermelli2020modeling, michieli2019incremental, michieli2021continual, douillard2021plop, cermelli2020few}.
% Performing incremental learning in object detection (ILOD) introduces additional challenges since every image contains multiple objects that may belong either to known classes or to classes to be learned. \cite{shmelkov2017incremental} defined ILOD considering that only the objects belonging to new classes are annotated, while the others are not considered.
A better solution is to use incremental learning and update the models directly to extend their knowledge to new classes by training only on new data and avoiding catastrophic forgetting \cite{mccloskey1989catastrophic}.
Incremental learning has primarily been studied in the context of image classification \cite{rebuffi2017icarl, douillard2020podnet, li2017learning, fini2020online, aljundi2018memory, rusu2016progressive, kirkpatrick2017overcoming} but it has only recently been applied to more complex tasks like object detection \cite{shmelkov2017incremental, peng2020faster, hao2019end, chen2019new, zhou2020lifelong} and semantic segmentation \cite{cermelli2020modeling, cermelli2021modeling, cermelli2022incremental, michieli2019incremental, michieli2021continual, douillard2021plop, cermelli2020few}.
Performing incremental learning in object detection (ILOD) poses additional challenges because each image contains multiple objects and, following the definition in \cite{shmelkov2017incremental}, only objects belonging to new classes are annotated while the rest (objects belonging to either old or future classes) are ignored, introducing missing annotations (see \cref{fig:teaser}).

Previous research has concentrated on introducing regularizations to prevent catastrophic forgetting, but the impact of missing annotations has been overlooked. Regions without annotations, in particular, are commonly considered as background areas, and the model assigns them to a special \textit{background} class. As a result, objects that are not annotated will be associated with the background, exacerbating catastrophic forgetting in old classes and making training more difficult in future classes.
%As a result, if an object from an older class does not have an annotation, it will be associated with the background, reducing performance and exacerbating catastrophic forgetting.
%\fabio{Qui va spiegato meglio perché - fai riferimento alla figura teaser e spiega che, quando il modello non vede annotazinoe in certa area, impara ad associarlo alla classe background.} 
% In this work, inspired by \cite{cermelli2020modeling}, we argue that considering non-annotated objects as background introduces the background shift and exacerbates the catastrophic forgetting while making harder to learn new classes.
% In particular, not taking into account that objects may be present in the background may induce the model to associate their regions to the special class \textit{background}, leading
% In particolare, considerare gli oggetti non annotati come background fa si che il modello impari ad associare questi oggetti alla speciale classe "background" e forzi il modello a dimenticare a quale classe appartenessero, introducing an issue similar to the background shift \cite{cermelli2020modeling} found in semantic segmentation.

To overcome this issue, inspired by \cite{cermelli2020modeling}, we revisit the common knowledge distillation framework in ILOD \cite{shmelkov2017incremental, peng2020faster, zhou2020lifelong} proposing MMA, that \textbf{M}odels the \textbf{M}issing \textbf{A}nnotations in both the classification and distillation losses. 
% In particular, on the classification loss, since forcing the model to predict the background on any region not associated to an annotation would exacerbate catastrophic forgetting, we flexibly allow it to predict either an old class or the background. Differently, since current classes may have been seen in previous learning step annotated as background, we revisit the distillation loss, matching the background probability of the teacher model with the probability of having either a new class or the background, allow to learn more easily the new classes.
% We demonstrate the benefit of our method on the Pascal-VOC 2007 \cite{pascal-voc-2007} dataset, considering multiple single-step and multi-step tasks. We show that our method is able to outperform the current state-of-the-art, without using any image from the past training steps.
We flexibly allow the model to predict either an old class or the background on any region not associated with an annotation on the classification loss to alleviate catastrophic forgetting. Alternatively, because current classes may have been annotated as background in a previous learning step, we revisit the distillation loss, matching the teacher model's background probability with the probability of having either a new class or the background, allowing new classes to be learned more easily.
On the Pascal-VOC dataset \cite{pascal-voc-2007}, we demonstrate the utility of our method by examining a variety of single-step and multi-step tasks. Without using any image from previous training steps, we show that our method outperforms the current state-of-the-art.

% Finally, we show that our framework can be easily extended to the instance segmentation task by introducing an additional knowledge distillation term. We evaluate our method on the Pascal SBD 2012 \cite{pascal-voc-2012} dataset, showing that it outperforms the other baselines.
Finally, we show that by adding an additional knowledge distillation term to our framework, we can easily extend it to the task of instance segmentation. On the Pascal SBD 2012 dataset \cite{pascal-voc-2012}, we show that our method outperforms the other baselines.

\noindent To summarize, the contributions of this paper are as follows:
\begin{itemize} 
    \itemsep0.1em 
    \item We identify the peculiar missing annotations issue in incremental learning for object detection.
    \item We propose to revisit the standard knowledge distillation framework to cope with the missing annotations, showing that our proposed MMA outperforms previous methods on multiple incremental settings.
    \item We extend our method to instance segmentation and we show that it outperforms all the other baselines.~\footnote{Code can be found here \url{https://github.com/fcdl94/MMA}.}
\end{itemize}

\section{Related work}
\label{sec:related}
\myparagraph{Object Detection.}
Object detection architectures can be mainly distinguished in two categories: one-stage detectors \cite{yolov3, tian2019fcos, liu2016ssd, carion2020end, tan2020efficientdet, zhou2019objects} and two-stage detectors \cite{girshick2014rich, girshick2015fast, ren2015faster, he2017mask, lin2017feature}.
Two-stage detectors are usually superior in performance but are less efficient, implementing two subsequent steps to perform detection: the model first extract regions of interest (RoIs) employing either a neural network \cite{ren2015faster} or an external region proposer \cite{girshick2015fast} and then use a MLP on the RoIs to obtain the final classification and bounding box regression.
Differently, one-stage detectors directly predict the final output, without requiring to predict RoIs.
These architectures are undoubtedly powerful in a standard, offline setting but they are not suited to incrementally add new classes over time. In this work, we focus on extending two-stage methods, in particular the Faster R-CNN \cite{ren2015faster}, to extend its knowledge on new categories without forgetting the previous knowledge in absence of the original data.

\myparagraph{Incremental Learning.}
The problem of catastrophic forgetting \cite{mccloskey1989catastrophic} has been extensively studied in the image classification task and recently extended to semantic segmentation. Previous works can be divided in three categories: rehearsal-based \cite{rebuffi2017icarl, castro2018end, shin2017continual, hou2019learning, wu2018memory, ostapenko2019learning}, regularization-based \cite{kirkpatrick2017overcoming,chaudhry2018riemannian,zenke2017continual,li2017learning, dhar2019learning} and parameter isolation-based \cite{mallya2018packnet, mallya2018piggyback, rusu2016progressive}. 
Rehearsal-based methods either store \cite{rebuffi2017icarl, castro2018end, hou2019learning, wu2019large} or generate \cite{shin2017continual, wu2018memory, ostapenko2019learning} examples of previous tasks, which are used to compensate for the lack of previous data during the training phase of the new task. 
Parameter isolation-based methods assign a subset of the parameters to each task and prevent them to change to avoid forgetting. Regularization-based methods can be divided in prior-focused and data-focused. 
The former \cite{zenke2017continual, chaudhry2018riemannian, kirkpatrick2017overcoming, aljundi2018memory} relies on knowledge stored in parameters value, constraining the learning of new tasks by penalizing changes of important old parameters. The latter \cite{li2017learning, dhar2019learning, fini2020online, douillard2020podnet, douillard2021dytox, yan2021dynamically, hou2019learning, cermelli2020modeling} exploits distillation \cite{hinton2015distilling} and uses the distance between the activation produced by the old network and the new one as a regularization term to prevent catastrophic forgetting. 
In this work, we focus on the data-focused regularization-based knowledge distillation approach by adapting it in the object detection context while modeling the missing annotations issue.
We note that \cite{cermelli2020modeling} identified a problem similar to the missing annotations in incremental semantic segmentation called background shift. We take inspiration from it to address the missing annotation problem in object detection.
% While image-level incremental learning was widely explored, in other vision tasks just few works exist and, especially in object detection, experimental protocols are fragmented. In this work we propose a distillation based framework with the aim of addressing the problem of modeling the background shift which is peculiar of the object detection task.

\myparagraph{Incremental Learning in Object Detection.}
Incremental learning in Object detection has witnessed more attention in last years. A pioneer work in this task is \cite{shmelkov2017incremental}, that proposes a framework based on two-stage object detectors by performing knowledge distillation on the output of Fast R-CNN \cite{girshick2015fast}. % without any rehearsal strategy.
Inspired by this work, some methods extend the distillation framework on the Faster R-CNN \cite{ren2015faster} architecture by adding distillation terms on the intermediate feature maps \cite{yang2021multiview, peng2020faster, chen2019new, liu2020multi, zhou2020lifelong} and proposing to further avoid forgetting on region proposal network \cite{peng2020faster, chen2019new, hao2019end, zhou2020lifelong}. Interestingly \cite{zhou2020lifelong} proposed a pseudo-positive-aware sampling algorithm to identify regions belonging to old classes and preventing them to be sampled as background regions. However, it only provides a partial solution for the missing annotation since it does not consider them in the distillation term nor the confidence of the model.  % considering pseudo-positive any region proposal having a probability for an old class greater than 0.5.
Other methods \cite{gupta2021ow, joseph2021towards, 2021ilodmeta, acharya2020rodeo} focused on rehearsal methods to maintain the old task knowledge, either performing replay of the intermediate features \cite{acharya2020rodeo} or the images \cite{li2019rilod, gupta2021ow, gupta2021ow}.
Differently, \cite{incdet} proposes a parameter isolated method extending EWC \cite{kirkpatrick2017overcoming} in the context of object detection. % The latter, in addition, exploit a dilatable classification head to alleviate forgetting.
Finally, a few works explored incremental learning utilizing one-stage architectures \cite{li2019rilod, peng2021diode, perez2020incremental}. 
In this work, we focus on proposing a distillation framework for two-stage architectures by explicitly modeling the missing annotations about object not belonging to the current training step.
% As for classic object detection also ICL pipelines show differences in architectures, in fact there is not a clear base learner to the aforementioned tasks, but they use both one or two stage architectures.   
% For the best of our knowledge just \cite{Gu_Deng_Wei_2021} worked on incremental instance segmentation using a YOLACT \cite{bolya2019yolact} and a multi-teacher distillation approach. 
% We note that PPAS \cite{zhou2020lifelong} proposed a pseudo-positive-aware sampling algorithm to avoid identifying past classes as negatives. However, they did not considered the model uncertainty and considered a pseudo-positive as any proposal having a probability for an old class greater than 0.5.

\section{Method}
\begin{figure*}
    \centering
    \includegraphics[width=0.95\textwidth]{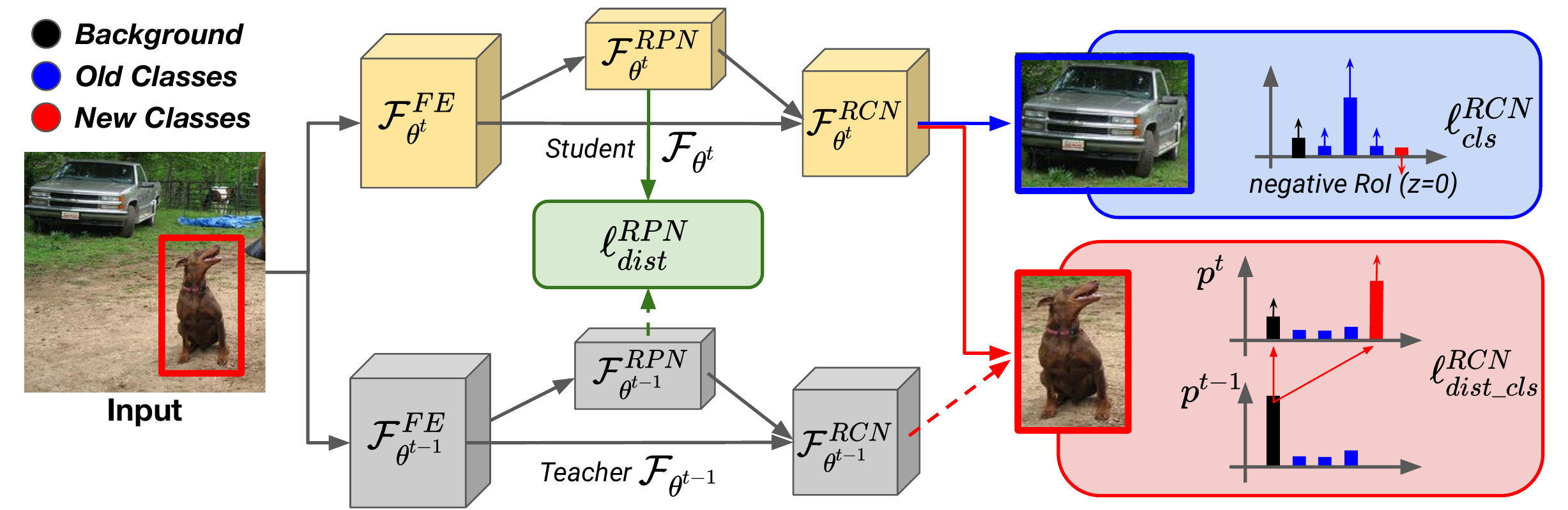}
    \caption{Overview of MMA, highlighting its contributions. Given an image, it is forwarded on the student (top) and teacher (bottom) models. The blue box illustrates the behavior of unbiased cross entropy loss on a negative RoI (\ie RoI without annotation): the model maximizes the probability of having either the background or an old class. In the red box, we show the effect of the unbiased distillation loss on the classification output for a new class region: it associates the teacher background with either the student background or a new class. Lastly, in green, it is reported the RPN distillation loss.}
    \label{fig:method}
    \vspace{-8pt}
\end{figure*}

\subsection{Problem Definition and Preliminaries} \label{sec:problem}
The goal of object detection is to train a model able to detect objects, \ie localize and classify them by producing a rectangular box and a class label. 
In this work, we focus on detection model in the R-CNN \cite{girshick2015fast, ren2015faster, he2017mask} family. 
A detection model, denoted $\detector_{\theta}$ with parameters $\theta$, is composed by three components: a feature extractor $\detector_{FE}$, a region proposal network (RPN) $\detector^{RPN}_{\theta}$, and a classification head $\detector_{\theta}^{RCN}$.
Denoting with $x$ an image, the feature extractor produces a dense feature map. % $\detector_{\theta}^{FE}(x) \in \real^{K \times H \times W}$, with $K$ indicating to the number of channels and $H$ and $W$ height and width of the image. 
The map is forwarded to the RPN with the goal of producing a set of $N$ rectangular regions of interest (RoIs), each associated with a binary objectness score. The $N$ RoIs are then applied to the feature map and classified by the classification head that produces, for each RoI, the class probabilities $p \in \real^{|\mathcal{C}|+1}$, indicating with $\mathcal{C}$ the set of classes, and the rectangular boxes $r \in \real^{4|\mathcal{C}|}$, one for each class. We note that the classifier also outputs a class score for the background to indicate that no objects are present in the RoI. %$\detector_{\theta}^{RCN}$ also outputs refined bounding box coordinates for each proposal.

In incremental learning for object detection (ILOD) the training is performed over multiple \textit{learning steps}, each one introducing a new set of classes to be detected. 
Formally, in the $t$-th training step, a detection model $\detector_{\theta^t}$ is updated to learn a set of classes $\mathcal{Y}^t$ employing a training set $\data^t$. % that provides annotations only for new classes. %$\{(y_i, \mathbf{r}_i)|i=1,\dots,M\}$ for each image $x$, where $M$ is the number of objects in $x$. 
We note that while an image in the training set $\data^t$ can contain multiple objects of different classes, following the ILOD protocol \cite{shmelkov2017incremental} \textit{only} annotations for classes in $\mathcal{Y}^t$ are provided. Moreover, at training step $t$ the old training sets are not available.
After the $t$-th step, the model $\detector_{\theta^t}$ is expected to produce prediction for all the classes seen so far, \ie its output should consider the classes in $\mathcal{C}^t = \cup_{t'=1}^{t} \mathcal{Y}^{t'}$. We note that $\mathcal{Y}^i \cap \mathcal{Y}^j = \emptyset$ for any $i,j \leq t$ and $ i \neq j$. 

\myparagraph{Faster R-CNN.}
In the standard Faster R-CNN \cite{ren2015faster} training is performed minimizing a multi-task loss as follows:
\begin{equation} \label{eq:faster}
    \ell_{faster} = \ell_{cls}^{RPN} + \ell_{reg}^{RPN} + \ell_{cls}^{RCN} +\ell_{reg}^{RCN}.
\end{equation}
The first two terms are the classification and regression loss on the RPN \cite{ren2015faster}, while the latter are applied on the classification head output \cite{girshick2015fast}.
Please refer to \cite{girshick2015fast, ren2015faster} for additional details on the training of Faster R-CNN.

\subsection{MMA: Modeling the Missing Annotations} \label{sec:mib}
% Despite being a powerful architecture, Faster R-CNN is not suited to update its weights to learn new classes. In particular, fine-tuning the model on $\data^t$ using \cref{eq:faster} induces the model to completely forget what it has learned, suffering {catastrophic forgetting} \cite{mccloskey1989catastrophic}.
% To alleviate it, previous works \cite{shmelkov2017incremental, peng2020faster, zhou2020lifelong, hao2019end, chen2019new} introduced knowledge distillation \cite{hinton2015distilling, li2017learning}: at the training step $t$, the \textit{student} model $\detector_{\theta^t}$ is forced to mimic the output of the \textit{teacher} model $\detector_{\theta^{t-1}}$, \ie the model frozen after the previous training step.

Despite its strength, Faster R-CNN is not well suited to updating its weights in order to learn new classes. Fine-tuning the model on $\data^t$ using \cref{eq:faster}, in particular, causes the model to forget everything it has learned, resulting in catastrophic forgetting \cite{mccloskey1989catastrophic}. To address this, previous research \cite{shmelkov2017incremental, peng2020faster, zhou2020lifelong, hao2019end, chen2019new} proposed the use of knowledge distillation \cite{hinton2015distilling, li2017learning}, in which, at the training step, the \textit{student} model $\detector_{\theta^t}$ is forced to mimic the output of the \textit{teacher} model $\detector_{\theta^{t-1}}$, \ie. the model at the previous training step.

Previous research, while addressing forgetting, did not address the issue of missing annotations. At time step $t$ the dataset $\data^t$ provides annotations only for objects in $\mathcal{Y}^t$ and other objects present in the image, belonging either to past or future classes, are not annotated. Following the standard detection pipeline, any RoI that does not match a ground truth annotation is associated to the background. This introduces two issues: (i) if the RoI contains an object of an old class, the model learns to predict it as background, exacerbating the forgetting; (ii) when the RoI contains an object that will be learned in the future, the model learns to consider it as background, making harder to learn new classes when presented.
%and associated, by the model, to a background region. 
%Thus, the background semantic changes at every time step and, without properly modeling it, it exacerbates forgetting and makes harder to learn new classes.
% Following \cite{cermelli2020modeling}, we show how to properly model the background in the context of object detection.
% - Recap of knowledge distillation \\
% - KD has been investigated by few-works in ILOD \\
% - However, they do not consider the background shift as in MiB. \\
% - In the following we will show how we adapt MiB to detection.
The missing annotation issue is similar to the background shift presented in \cite{cermelli2020modeling} in the context of incremental learning for semantic segmentation. In the following, we show how to adapt the equations proposed by \cite{cermelli2020modeling} in incremental learning for object detection.

\myparagraph{Unbiased Classification Loss.}
The classification loss $\ell_{cls}^{RCN}$ in the Faster R-CNN has the goal to force the network to produce the correct class label for the RoIs. In detail, given a sampled set of $N$ RoIs generated by the RPN and matched with a ground truth label (positive RoI) or with the background (negative RoI), the loss is computed as:
\begin{equation} \label{eq:faster_cls}
    \ell_{cls}^{RCN} = \frac{1}{N} \sum_{i=1}^{N} z_i (\sum_{c \in \mathcal{C}^t} \bar{y}_i^c \log(p_i^c))  + (1-z_i) \log(p_i^b),
\end{equation}
where $z_i$ is 1 for a positive RoI and 0 otherwise, $\bar{y}_i$ is the one-hot class label (1 for the ground truth class, 0 otherwise), and $p_i^b$ indicates the probability for the background class for the $i$-th RoI. 

The \cref{eq:faster_cls} does not consider that only information about novel classes is available in the ground truth because it was designed for standard object detection. The problem is that all other objects in the image that are not associated with any ground-truth are treated as a negative RoI and the model learns to predict the background class on them.
This problem is especially harmful for objects of old classes because it causes the model to forget the object's correct class and replace it with the background class, resulting in severe catastrophic forgetting.
% At time $t$, objects of past classes $\mathcal{C}^{t-1}$ that appear in $\data^t$ are not annotated and any RoI matching their region is considered negative, \ie it is assumed it is a background region. For any negative RoI, in \cref{eq:faster_cls}  is forced to to predict the background class for
%\cref{eq:faster_cls} forces the network to produce predictions the ground-truth labels on positive proposals, and the background on negative proposal. However, since we do not have annotation for old classes in the current training step, proposal that actually contains an old class are considered negative and thus predicted as background, introducing the background shift issue \cite{cermelli2020modeling} and exacerbating catastrophic forgetting.

To avoid this issue, we modify \cref{eq:faster_cls} as follows:
\begin{equation} \label{eq:uce}
\begin{aligned}
    \ell_{cls}^{RCN} = \frac{1}{N} \sum_{i=1}^{N} z_i (\sum_{c\in \mathcal{Y}^t} \bar{y}_i^c \log(p_i^c)) + \\ (1-z_i) \log(p_i^b + \sum_{o\in \mathcal{C}^{t-1}} p_i^o),
\end{aligned}    
\end{equation}
where $p_i^c$ is the probability of class $c$ for query $i$,  $\mathcal{Y}^t$ are the new classes at $t$ and $\mathcal{C}^{t-1}$ are all the classes seen before $t$.
Using \cref{eq:uce} the model learns new classes on the positive RoIs ($z_i = 1$) while preventing the background to supersede the old classes: instead of forcing the background class on every negative RoI ($z_i = 0$), as in \cref{eq:faster_cls}, it forces the model to predict either the background or any old class by maximizing the sum of their probabilities.
An illustration is reported in the blue box of \cref{fig:method}.

% We note that PPAS \cite{zhou2020lifelong} proposed a pseudo-positive-aware sampling algorithm to avoid identifying past classes as negatives. However, they did not considered the model uncertainty and considered a pseudo-positive as any proposal having a probability for an old class greater than 0.5.
%Using Eq.~Y, we maintain the Eq.~X for the positive proposals, while avoiding the background shift. In particular, instead of forcing the background class, we maximize the probability of having either the background or an old class on a negative proposal.

\myparagraph{Unbiased Knowledge Distillation.}
A common solution to avoid forgetting is to add two knowledge distillation loss terms to the training objective  \cite{peng2020faster, zhou2020lifelong, hao2019end, chen2019new}:
\begin{equation} \label{eq:dist}
\begin{aligned}
    \ell = \ell_{faster} + \lambda^{1} \ell_{dist}^{RCN} + \lambda^{2} \ell_{dist}^{RPN},
\end{aligned}
\end{equation}
where $\lambda^{1}$, $\lambda^{2}$ are hyper-parameters.

The goal of $\ell_{dist}^{RCN}$ is to maintaining the knowledge about old classes on the classification head. Previous works \cite{shmelkov2017incremental, peng2020faster} force the student model to output classification scores and box coordinates for old classes close to the teacher employing an L2 loss.  %, considering only the old classes and the background. 
However, they ignore the missing annotations, \ie the new classes have been observed in previous steps but, since they had been observed without annotations, they have been associated to the background class.
The teacher would predict an high background score for new classes RoIs, and forcing the student to mimic its behavior would make harder to learn new classes, contrasting the classification loss.
%Without considering this issue, the student is forced to predict background on new class regions, contrasting the loss in \cref{eq:uce} and making harder learn new classes.
Taking model the missing annotations, we formulate the distillation loss as:
\begin{equation} \label{eq:dist_rcn}
    \ell_{dist}^{RCN} = \frac{1}{N} \sum_{i=1}^{N} \ell_{dist\_cls}^{RCN}(i) + \ell_{smooth\_l1}(r_i^t, r_i^{t-1}),
\end{equation}
\begin{equation} \label{eq:ukd}
\begin{aligned}
    \ell_{dist\_cls}^{RCN}(i) = \frac{1}{|\mathcal{C}^{t-1}|+1} (p_{i}^{b, t-1} \log(p_{i}^{b,t} + \sum_{j\in \mathcal{Y}^{t}} p_{i}^{j, t}) \\ + \sum_{c\in\mathcal{C}^{t-1}} p_{i}^{c, t-1} \log(p_{i}^{c,t})),
\end{aligned}
\end{equation}
where $p_{i}^{k, t-1}$, $r_i^{t-1}$ and $p_{i}^{k,t}$, $r_i^t$ indicates, respectively, the classification and regression output for the proposal $i$ and class $k$ of the teacher and student model, and $b$ is the background class.
While the second term of \cref{eq:dist_rcn} has been used in previous works \cite{shmelkov2017incremental, peng2020faster} and considers the box coordinates, we propose to modify the first term that is responsible to handle the classification scores.
To model the missing annotations, \cref{eq:ukd} uses the all the class probabilities of the student model to match the teacher ones: the old classes $\mathcal{C}^{t-1}$ are kept unaltered among student and teacher models, while the background of the teacher $p_i^{b, t-1}$ is associated with either a new class or the background in the student. 
With \cref{eq:ukd}, when the teacher predicts an high background probability for a RoI belonging to a new class, the student is not forced to mimic its behavior but it can consolidate its new knowledge and predict the correct class. An illustration is reported in the red box of \cref{fig:method}.

On the other hand, $\ell_{dist}^{RPN}$ goal is to avoid forgetting on the RPN output. Since annotation for old classes are not available, the RPN learns to predict an high objectness score only on RoIs belonging to new classes. To force the RPN to 
%operates on the RPN output and aims to force the model to 
maintain an high objectness score for regions belonging to old classes, we use the loss proposed by \cite{peng2020faster}. The student is forced to mimic the teacher only on regions belonging to old classes, \ie where the teacher score is greater than the student one. %, \ie when the proposal belongs to either the background or an old class, while ignoring proposals of new classes.
Considering A regions, we compute $\ell_{dist}^{RPN}$ as:
% and we compute $\ell_{dist}^{RPN}$ on A object anchors as follows:
\begin{equation} \label{eq:dist_rpn}
\begin{aligned}
    \ell_{dist}^{RPN} = \frac{1}{A} \sum_{i = 1}^{A} \mathds{1}_{[s^t_i \geq s^{t-1}_i]} ||s^t_i - s^{t-1}_i|| + \\ \mathds{1}_{[s^t_i \geq s^{t-1}_i + \tau]} ||\omega^t_i - \omega^{t-1}_i||,
\end{aligned}
\end{equation}
where $s^t_i$ is the objectness score and $\omega^t_i$ the coordinates of $\detector^{RPN}_{\theta^t}$ on the $i$-th proposal, $||\cdot||$ is the euclidean distance, $\tau$ is an hyperparameter, and $\mathds{1}$ is the indicator function equal to 1 when the condition on the brackets is verified and 0 otherwise.
Note that when $s^t_i > s^{t-1}_i$, the teacher produces an objectness score greater then the student and the proposal is probably containing an old class. Differently, when $s^t_i \geq s^{t-1}_i$, the proposal is likely belonging to a new class and forcing the student to mimic the teacher score may introduce errors that hamper the performance on new classes.

\subsection{Extension to Instance Segmentation}
The goal of instance segmentation is to produce a precise pixel-wise mask for each object in the image. % instead of a coarse bounding box. 
To produce masks we rely on Mask R-CNN \cite{he2017mask}, that extends the Faster R-CNN introducing a mask head $\detector^{MASK}_{\theta}$. It produces, for each RoI, an additional binary segmentation mask with shape ${|\mathcal{C}| \times h \times w}$, where $\mathcal{C}$ is the set of classes and $h$,$w$ is the mask resolution.
To train the mask head, \cite{he2017mask} introduces an additional loss term that is summed to the multi-task loss in \cref{eq:faster}. Formally, Mask R-CNN objective is:
\begin{equation} \label{eq:mask}
    \ell_{mask} = \ell_{faster} + \ell^{MASK}_{cls},
\end{equation}
where $\ell^{MASK}_{cls}$ is the per-pixel binary cross-entropy loss between the $\detector^{MASK}_{\theta}$ output and the binary mask of the ground truth class. 
Please refer to \cite{he2017mask} for details.

Despite the method presented in \cref{sec:mib} already accounts for forgetting on the detection head, by applying \cref{eq:mask} we incur the risk to forget how to segment past objects while learning the new ones.
For this reason, we further extend \cref{eq:dist} to add a knowledge distillation term on the mask head.
Formally, in instance segmentation we employ the following training objective:
\begin{equation} \label{eq:dist2}
\begin{aligned}
    \ell = \ell_{mask} + \lambda_{1} \ell_{dist}^{RCN} + \lambda_{2} \ell_{dist}^{RPN} + \lambda_{3} \ell_{dist}^{MASK},
\end{aligned}
\end{equation}
where $\lambda_1$, $\lambda_2$, $\lambda_3$ are hyper-parameters.

$\ell_{dist}^{MASK}$ has the goal of keeping the segmentation mask for old classes close to the output of the teacher model. In particular, we employ a per-pixel binary cross-entropy loss between the teacher model masks and the student ones. Formally, denoting as $m^t_{c,i}$ the segmentation mask produced by $\detector^{MASK}_{\theta^t}$ for the class $c$ at pixel $i$, we compute
\begin{equation} \label{eq:dist_mask}
\begin{aligned}
    \ell_{dist}^{MASK} = \frac{1}{|I| |\mathcal{C}^{t-1}|} \sum_{i \in I} \sum_{c \in \mathcal{C}^{t-1}} m^{t-1}_{c,i} \log(m^{t}_{c,i}) + \\ (1 - m^{t-1}_{c,i}) \log(1 - m^{t}_{c,i}),
\end{aligned}
\end{equation}
where I is the set of pixels and $|I| = h \times w$.
We note that \cref{eq:dist_mask} is computed only on the segmentation masks belonging to old classes in $\mathcal{C}^{t-1}$, while the masks belonging to the new ones are not considered.

\section{Experiments}
\label{sec:experiments}

\begin{table*}[t]
\centering
\caption{mAP@0.5\% results on single incremental step on Pascal-VOC 2007. Methods with $\dagger$ come from reimplementation. Methods with * use exemplars.}
\setlength{\tabcolsep}{7pt}
\begin{adjustbox}{width=\linewidth}
\begin{tabular}{l|cccc|cccc|cccc}
\toprule
& \multicolumn{4}{c|}{\textbf{19-1}}  & \multicolumn{4}{c|}{\textbf{15-5}}   & \multicolumn{4}{c}{\textbf{10-10}} \\ 
\textbf{Method} & \textbf{1-19}    & \textbf{20}    & \textbf{1-20}    & \textbf{Avg}     & \textbf{1-15}    & \textbf{16-20}    & \textbf{1-20}    & \textbf{Avg}   & \textbf{1-10}    & \textbf{11-20}    & \textbf{1-20}    & \textbf{Avg}  \\ \hline
Joint Training		        & 75.3 & 	73.6 &	75.2 &	74.4 &	76.8 &	70.4&	75.2&	73.6	& 74.7&     75.7 &	75.2 &	75.2 \\\hline
% Base		        & 72.1 & 		 &       &       &  75.5 &      &       &           & 76.5 &    	 &  	 &	     \\	\hline
Fine-tuning	                & 12.0 & 	62.8 &	14.5 &	37.4 &	14.2 &	59.2&	25.4&	36.7	&  9.5 &	62.5 &	36.0 &	36.0 \\	
ILOD (Fast R-CNN) \cite{shmelkov2017incremental}		        & 68.5 & 	62.7 &	68.3 &	65.6 &	68.3 &	58.4&	65.9&	63.4	& 63.2 &	63.1 &	63.2 &	63.2 \\
ILOD (Faster R-CNN) \cite{shmelkov2017incremental} $\dagger$ 	& 70.3 & 	65.2 &	70.0 &	\textbf{67.8} &	72.5 &	58.0&	68.9&	65.3	& 69.2 &	53.0 &	61.1 &	61.1 \\
Faster ILOD	\cite{peng2020faster}	    & 68.9 & 	61.1 &	68.5 &	65.0 &	71.6 &	56.9&	67.9&	64.3	& 69.8 &	54.5 &	62.1 &	62.1 \\
Faster ILOD \cite{peng2020faster} $\dagger$      & 70.9 & 	64.3 &	70.6 &	67.6 &	73.5 &	55.6&	69.1&	64.6	& 71.1 &	52.3 &	61.7 &	61.7 \\
PPAS \cite{PPAS}    & 70.5 & 	53.0 &	69.2 &	61.8 &	     &	    &        &          & 63.5 &	60.0 &	61.8 &	61.8 \\
MVC	\cite{yang2021multiview}	            & 70.2 & 	60.6 &	69.7 &	65.4 &	69.4 &	57.9&	66.5&	63.7	& 66.2 &	66.0 &	66.1 &	66.1 \\ 	\hline
OREO* \cite{joseph2021towards} 		        & 69.4 & 	60.1 &	68.9 &	64.7 &	71.8 &	58.7&	68.5&	65.2	& 60.4 &	68.8 &	64.6 &	64.6 \\
OW-DETR* \cite{gupta2021ow}		    & 70.2 & 	62.0 &	69.8 &	66.1 &	72.2 &	59.8&	69.1&	66.0	& 63.5 &	67.9 &	65.7 &	65.7 \\
ILOD-Meta* \cite{2021ilodmeta}	    & 70.9 & 	57.6 &	70.2 &	64.2 &	71.7 &	55.9&	67.8&	63.8	& 68.4 &	64.3 &	66.3 &	66.3 \\ \hline
\textbf{MMA}	            & 71.1 & 	63.4 &	\textbf{70.7} &	67.2 &	73.0 &	60.5&	\textbf{69.9}&	\textbf{66.7}	& 69.3 &	63.9 &	\textbf{66.6} &	\textbf{66.6} \\     \toprule
\end{tabular}
\end{adjustbox}
\label{tab:voc_exps_ss}
\end{table*}

\begin{table*}[t]
\centering
\caption{mAP@0.5\% results on multi incremental steps on Pascal-VOC 2007. Methods with $\dagger$ come from reimplementation.}
\setlength{\tabcolsep}{3pt}
\label{tab:voc_exps_ms}
\begin{adjustbox}{width=\linewidth}
\begin{tabular}{l|cccc|cccc|cccc|cccc}
\hline
 & \multicolumn{4}{c|}{\textbf{10-5}}   & \multicolumn{4}{c|}{\textbf{10-2}} & \multicolumn{4}{c|}{\textbf{15-1}} & \multicolumn{4}{c}{\textbf{10-1}} \\
\textbf{Method}                                                             & \textbf{1-10} & \textbf{11-20} & \textbf{1-20}          & \textbf{Avg-S} & \textbf{1-10}  & \textbf{11-20}   & \textbf{1-20}   & \textbf{Avg-S}  & \textbf{1-15}   & \textbf{16-20}   & \textbf{1-20}   & \textbf{Avg-S}  & \textbf{1-10}   & \textbf{11-20}   & \textbf{1-20} & \textbf{Avg-S}  \\ \hline
Joint Training                                                      & 74.7 & 75.7  & 75.2          & 75.2  & 74.7   & 75.7    & 75.2   & 75.2   & 76.8   & 70.4    & 75.2   & 73.5   & 74.7   & 75.7    & 75.2  & 75.2   \\ \hline
% Base                                                         & 76.5 &       &               &       & 76.5   &         &        &        & 75.5   &         &        &        & 76.5   &         &       &        \\ \hline
Fine-tuning                                         & 6.6  & 28.3  & 17.4          & 21.8  & 5.2    & 12.3    & 8.8    & 16.7   & 0.0    & 8.0     & 2.4    & 6.7    & 0.0    & 4.6     & 2.3   & 8.6    \\
ILOD (Faster R-CNN) \cite{shmelkov2017incremental} $\dagger$ & 67.2 & 59.4  & 63.3          & 65.2  & 62.1   & 49.8    & 55.9   & 62.2   & 65.6   & 47.6    & 60.2   & 65.8   & 52.9   & 41.5    & 47.2  & 59.1   \\
Faster ILOD \cite{peng2020faster} $\dagger$                  & 68.3 & 57.9  & 63.1          & 65.5  & 64.2   & 48.6    & 56.4   & 62.8   & 66.9   & 44.5    & 61.3   & 67.1   & 53.5   & 41.0    & 47.3  & 60.4   \\ \hline
\textbf{MMA}                                                         & 66.7 & 61.8  & \textbf{64.2} & \textbf{67.3}  & 65.0   & 53.1    & \textbf{59.1}   & \textbf{63.8}   & 68.3   & 54.3    & \textbf{64.1 }  & \textbf{67.5}   & 59.2   & 48.3    & \textbf{53.8}  & \textbf{62.4}   \\ \hline
\end{tabular}
\end{adjustbox}
\end{table*}

\subsection{Experimental Protocol}
We evaluate MMA on the Pascal-VOC dataset. In particular, following previous works, we employ PASCAL-VOC 2007 \cite{pascal-voc-2007} for object detection. It is a widely used benchmark that includes 20 foreground object classes and consists in 5K images for training and 5K for testing. For instance segmentation, we employed Pascal SBD 2012 \cite{BharathICCV2011}, that contains the same set of 20 classes but also reports the instance segmentation annotations. We used the standard split of Pascal SBD 2012, using 8498 images for training and 2857 for evaluation.
Following \cite{shmelkov2017incremental}, for both object detection and instance segmentation we implement the following experimental protocol: each training step contains all the images that have at least one bounding box of a novel class. % with only the latter annotated. 
We remark that at each training step it is assumed to have only labels for bounding boxes of novel classes, while all the other objects that appear in the image, either belonging to past or future classes, are not annotated. 
% the old ones are not labeled as background in the ground truth, each training step  contains all the images that have at least one bounding box of a novel class, with only the latter annotated.
This is a very realistic setup since it does not make any assumption on the objects present in the images and reduces the amount of annotation required in each incremental step.
% It is crucial to note that: images may contain objects of classes that we will learn in the future, but labeled as background.

\subsection{Implementation Details}
For object detection, we followed previous works \cite{peng2020faster, zhou2020lifelong, yang2021multiview, gupta2021ow, joseph2021towards, 2021ilodmeta} and we use the Faster R-CNN architecture with a ResNet-50 backbone. Similarly, for instance segmentation, we employ the Mask R-CNN \cite{he2017mask} architecture with ResNet-50 backbone. Both backbones are initialized using the ImageNet pretrained model \cite{deng2009imagenet}.
We used the same training protocol of \cite{shmelkov2017incremental, peng2020faster} but we increased the batch size from 1 to 4 to reduce the time required for training, scaling accordingly the learning rate and number of iterations. In particular, for object detection we train the network with SGD, weight decay $10^{-4}$ and momentum 0.9. We use an initial learning rate of $4 \cdot 10^{-3}$ for the first learning step and $4 \cdot 10^{-4}$ in the followings. We performed 10K iterations when adding 5 or 10 classes, while we trained for 2.5K when learning only one or two classes. We apply the same data augmentation of \cite{shmelkov2017incremental, peng2020faster}.
We set $\lambda_2$ equal to $0.1$, $0.5$, and $1$ when adding 10 classes, 5, and 1 or 2 classes, respectively. $\lambda_1$, $\lambda_3$ are set to $1$.

\subsection{Object Detection Results}
As done by previous works \cite{yang2021multiview, zhou2020lifelong, 2021ilodmeta, shmelkov2017incremental, peng2020faster}, for incremental object detection we evaluate our method considering experimental settings adding a different number of classes in one or multiple training steps. We report adding 10 (\textit{10-10}), 5 (\textit{15-5}) or 1 (\textit{19-1}) class in a single incremental step and performing two incremental steps adding 5 classes (\textit{10-5}), five steps adding two classes (\textit{10-2}) and either ten (\textit{10-1}) or five (\textit{15-1}) steps adding one class. As in previous works, we split the classes following the alphabetical order.

\myparagraph{Single-step incremental settings (10-10, 15-5, 19-1).}
% In this experiments, we perform two learning steps: the first in which we observe the first K classes, and the second where we learn the last $20-K$ remaining classes.
Results are reported in \cref{tab:voc_exps_ss}. The \textit{Avg} metric equally weights new and old classes averaging their aggregated mAP.
We benchmark MMA against previous works reporting the results on the same settings. We compare either with methods using rehearsal \cite{joseph2021towards, gupta2021ow, 2021ilodmeta} or not using them \cite{yang2021multiview, zhou2020lifelong, shmelkov2017incremental, peng2020faster}. We underline that the former methods are not compared fairly with MMA, since we do not use any replay memory to store old samples. Furthermore, for a fair comparison we report ILOD \cite{shmelkov2017incremental} and Faster ILOD \cite{peng2020faster} using our same architecture and training protocol. Finally, we report two simple baselines: the joint training upper bound, where the architecture is trained using the whole dataset and all the annotations, and the fine-tuning, where the architecture is trained on the new data using \cref{eq:faster}, without employing any regularization strategy.

As can be noted in \cref{tab:voc_exps_ss}, fine-tuning suffers a large drop in performance on the old classes, clearly indicating that catastrophic forgetting is an issue to be addressed.
While previous works improve the performance, addressing the forgetting issue, MMA outperforms all the previous methods, also the ones that uses exemplars to avoid forgetting, demonstrating the validity of our approach.
In particular, when comparing with ILOD \cite{shmelkov2017incremental} and Faster ILOD \cite{peng2020faster}, we note that our method achieve comparable performance on old classes but outperforms them on the new classes, outperforming them of 1\% on both 19-1 and 15-5, and even by 10\% on the 10-10 setting. We argue that the improvement is largely due to the unbiased distillation loss, that modeling the missing annotations, removes incoherent training objectives, increasing the performance. 
Comparing MMA to previous state-of-the-art, we note that it outperforms the competitive rehearsal strategies in every setting. On the 19-1 setting, MMA outperforms the ILOD-Meta by 0.5\% considering equally every class (\textit{1-20}) and by 1.1\% OW-DETR when considering equally old and new classes (\textit{Avg}). Similarly, in the 15-5 and 10-10 settings, MMA outperforms the best rehearsal method by 0.9\% and 0.3\% on all the classes 0.7\% and by 0.3\% on the \textit{Avg} metric, respectively.

% In both settings our method outperforms strategies that do not use exemplars and also some of the strategies with rehearsal, demonstrating the validity of the approach without using any additional memory.\\
% It is important to notice that our method boosts the performances on new classes significantly. In fact one of the major issues in object detection ICL is that it that does not suffer badly of catastrophic forgetting, on the other hand the model starts being exhausted after the first learning step, therefore new classes are very challenging to be learned. In fact looking at \cite{shmelkov2017incremental}, the less constrained ICL strategy, it obtain the highest performances on new classes, degrading more in the older ones. Our strategy succeeds in  counteracting the effect of a constrictive distillation.

\begin{figure*}[t]
\centering
\subfloat{
  \includegraphics[width=0.45\linewidth]{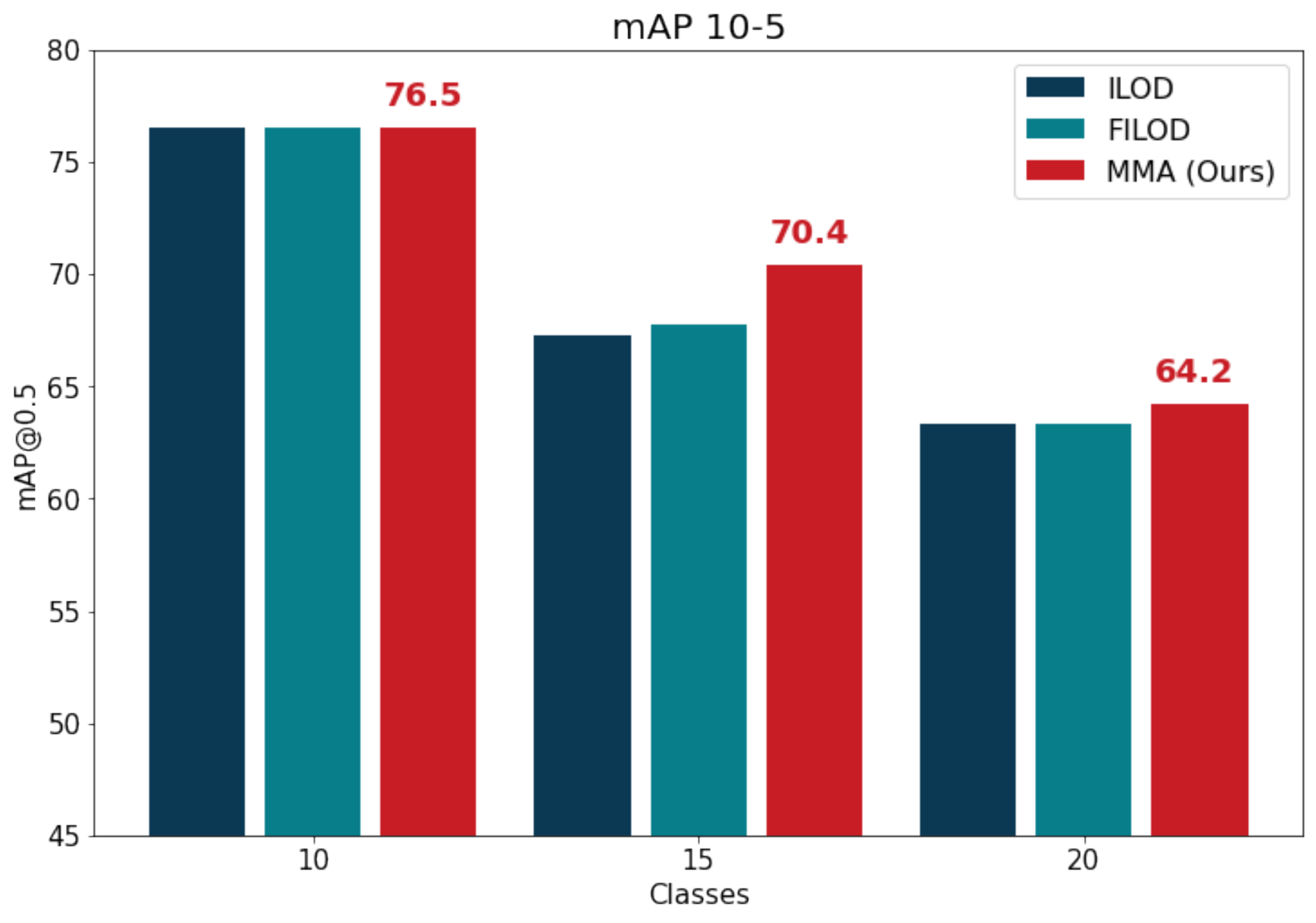}
}
\subfloat{
  \includegraphics[width=0.45\linewidth]{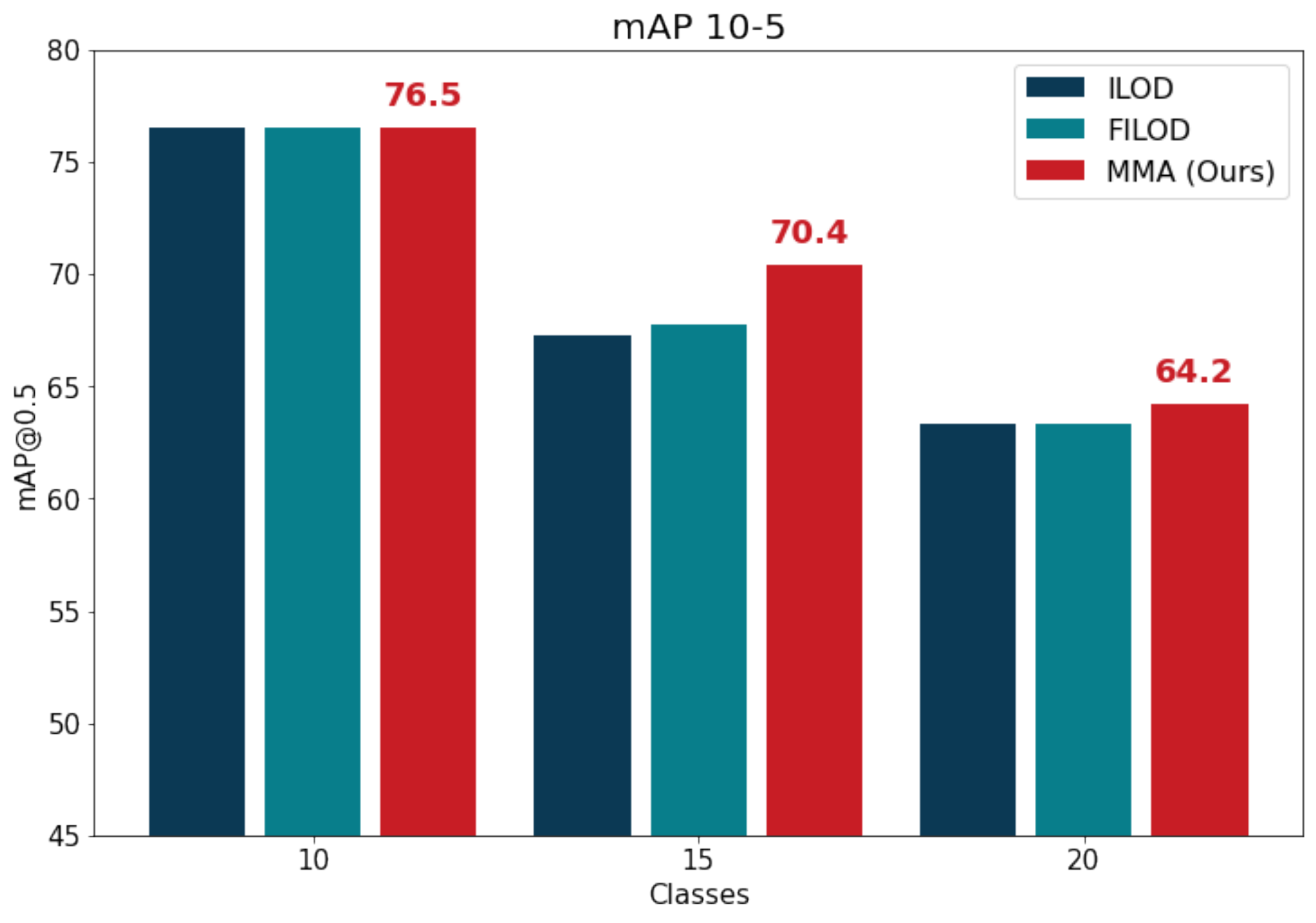}
}
\hspace{0mm}
\subfloat{
  \includegraphics[width=0.45\linewidth]{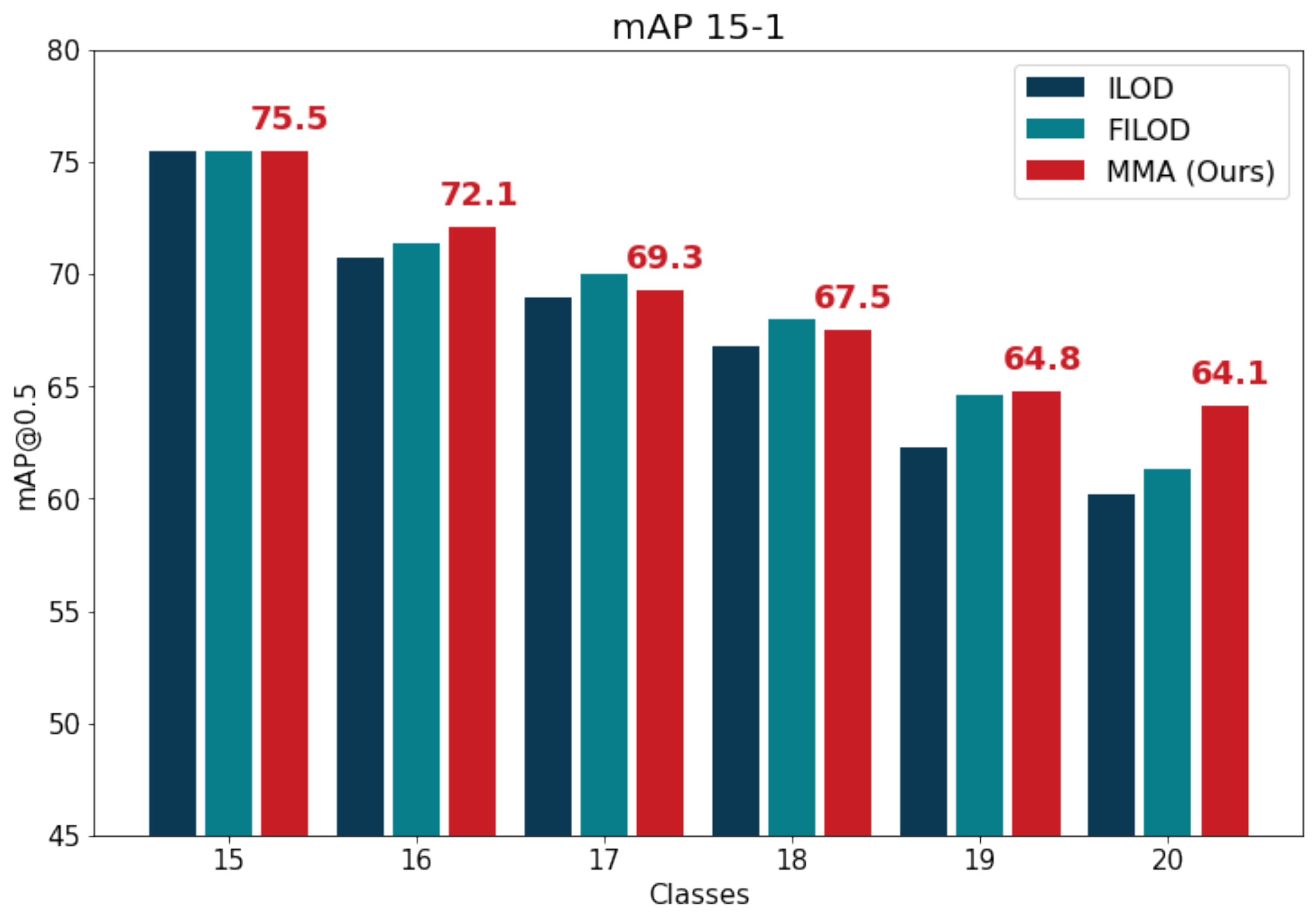}
}
\subfloat{
  \includegraphics[width=0.45\linewidth]{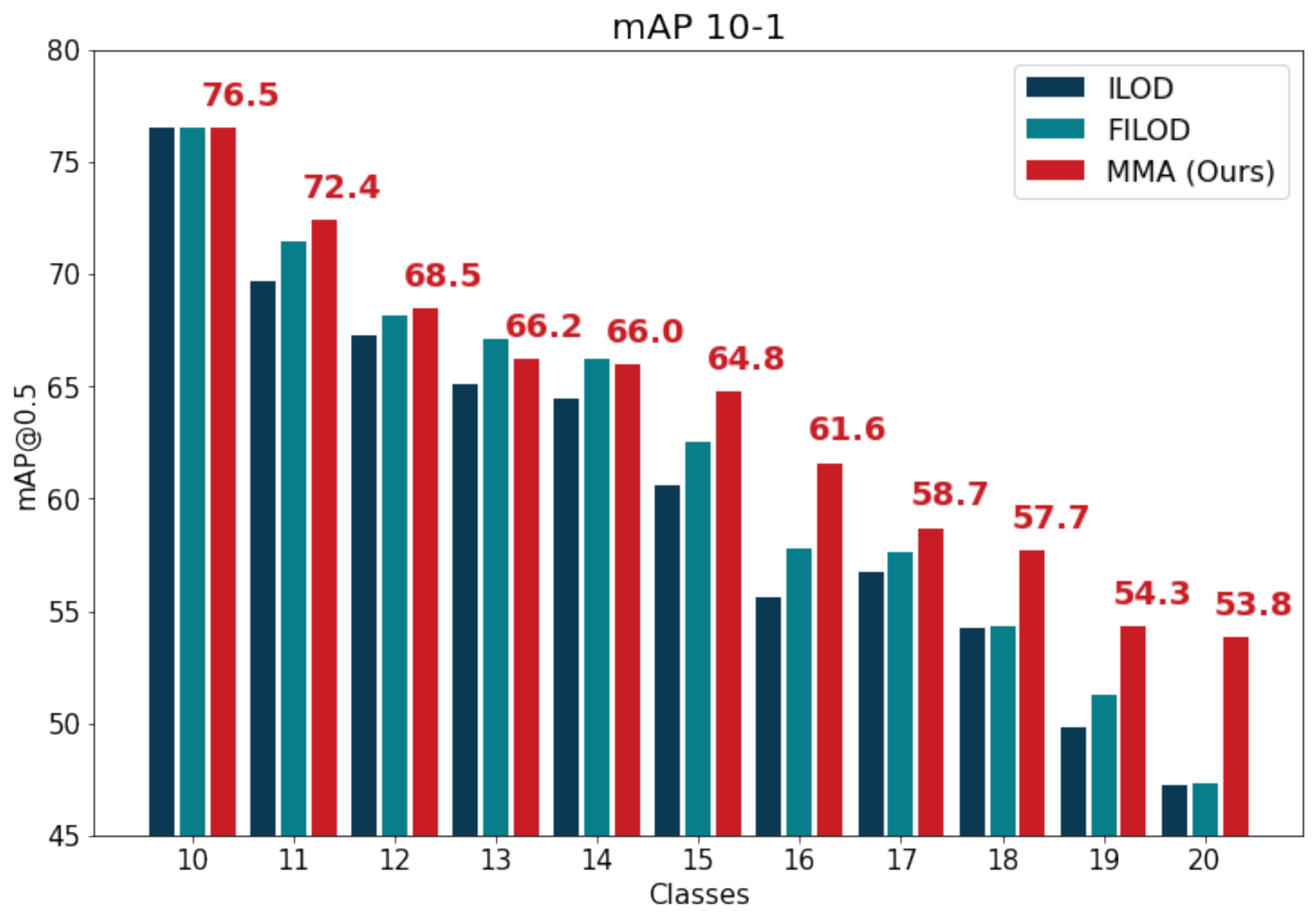}
}
\caption{mAP\% results on multiple incremental steps on Pascal-VOC 2007.}
\label{fig:multiStep}
\end{figure*}

\myparagraph{Multi-step incremental settings (10-5, 10-2, 15-1, 10-1).}
While performing a single training step is valuable to evaluate the ability to alleviate catastrophic forgetting, a more realistic setting is to perform multiple incremental steps adding new classes. In this section, we analyze the behavior of MMA against three baselines: fine-tuning, ILOD \cite{shmelkov2017incremental}, Faster ILOD \cite{peng2020faster}, all implemented following our experimental protocol. We report the results for the four considered settings in \cref{tab:voc_exps_ms}, showing the mAP\% over multiple incremental steps and \cref{fig:multiStep}, where the results after the last incremental step are reported. \cref{tab:voc_exps_ms} further reports the average performance across multiple steps \textit{Avg-S}.

We can observe that performing multiple incremental steps is challenging and existing methods performances drop badly compared to single step scenarios. In particular, fine-tuning the network on new data, without using any technique to avoid forgetting, lead to completely forgets the old classes, reaching, in the last step, performances close to 0\% on old classes. 
ILOD \cite{shmelkov2017incremental} and Faster ILOD \cite{peng2020faster} substantially alleviate catastrophic forgetting, leading to better results both on old and new classes. 
However, when comparing with MMA, we see that both ILOD and Faster ILOD achieve worse results. In particular, after the last step, it is evident that MMA obtain better performances on novel classes: +2.4\% o 10-5, +3.3\% on 10-2, +6.3\% on 15-1, and 6.8\% on 10-1 \wrt the best among the baselines. Furthermore, MMA also obtains comparable or greater performance than previous methods on the old classes.
Overall, MMA outperforms the best among ILOD and Faster ILOD by 0.9\% on 10-5, 2.7\% on 10-2, 2.8\% on 15-1, and 6.5\% on the 10-1 setting. We note that the improvement is larger when adding more classes, indicating that our method is better suited to performing multiple-incremental steps.
Considering the trend over multiple training steps in \cref{fig:multiStep}, we note that MMA is always comparable or better than previous methods. In particular, it is remarkable that MMA largely outperforms the other methods when increasing the number of training steps, as shown in the 10-1 setting.
% are decreasing their score especially on new classes, in fact especially in this setting, increasing the number of incremental steps the model loose the capability to learn new classes. On the other hand, our method is able to achieve better performances.
% Compared to the other approaches, our methods outperforms all baselines, increasing slightly the performances on old classes or  maintaining them equal while consistently improve results on new ones.
% Our method behave better than baselines with the growth of the number of steps, in fact the 10-1 scenario, in which ten classes are added one by one, demonstrate the highest GAP in Avg-MAP (+7\%), secondly in 10-2 (+4\%).\\ This behavior demonstrate that tackling the problem of background allows the model to not confusing new classes with background along incremental steps, allowing to learn also along many additions of new classes.

\begin{table}[t]
\centering
\caption{mAP@(0.5,0.95)\% results of incremental instance segmentation on Pascal-VOC 2012.} \label{tab:inst}
\setlength{\tabcolsep}{4pt}
\begin{adjustbox}{width=\linewidth}
\begin{tabular}{l|cccc|cccc} \toprule
                       & \multicolumn{4}{c|}{\textbf{19-1}}   & \multicolumn{4}{c}{\textbf{15-5}}           \\
\textbf{Method}        & \textbf{1-19}  & \textbf{20}   & \textbf{1-20}          & \textbf{Avg}  & \textbf{1-15}          & \textbf{16-20}         & \textbf{1-20} & \textbf{Avg}  \\ \hline
Joint Training                & 40.4  & 54.1 & 41.1          & 47.2 & 41.0          & 41.2          & 41.1 & 41.1 \\ \hline
% Base                   & 41.7 &      &               &      & 42.6          &               &      &      \\ \hline
Fine-tuning                     & 6.7   & 46.3 & 8.7           & 26.5 & 1.9           & 35.3          & 10.2 & 18.6 \\
Fine-tuning w/ \cref{eq:uce}                 & 12.5  & 47.5 & 14.3          & 30.0 & 13.0          & 35.5          & 18.6 & 24.2 \\
ILOD \cite{shmelkov2017incremental}                 & 40.1  & 38.3 & 40.0          & 39.2 & 39.2          & 30.8          & 37.1 & 35.0 \\
Faster ILOD \cite{peng2020faster}              & 40.6  & 38.1 & {40.4} & 39.3 & {39.4} & {30.3} & 37.1 & 34.8 \\ \hline
\textbf{MMA}            & 40.6  & 43.0 & 40.8          & 41.8 & 38.2          & 33.7          & 37.1 & 35.9 \\
\textbf{MMA + $\ell^{MASK}_{dist}$}      & 41.0  & 42.8 & \textbf{41.1}          & \textbf{41.9} & 40.2          & 32.2          & \textbf{38.2} & \textbf{36.2} \\ \toprule
\end{tabular}
\end{adjustbox}
\end{table}

\subsection{Instance Segmentation Results}
Following the protocol used in incremental object detection, we evaluate our method considering two experimental settings: adding one (\textit{19-1}) and five (\textit{15-5})classes in a single training step. As in object detection, we follow the alphabetical order of the dataset. Following the standard practice on instance segmentation, we report the mAP averaged across 11 IoU thresholds, ranging from 0.5 to 0.95, with a step of 0.05.
We compare MMA with fine-tuning, fine-tuning using the unbiased classification loss (\cref{eq:uce}), ILOD \cite{shmelkov2017incremental} and Faster ILOD \cite{peng2020faster}. For all the methods we employ the same architecture and hyper-parameters.

\cref{tab:inst} shows the results for the 19-1 and 15-5 settings, reporting the average mAP of new and old classes separately, the average over all classes, and the average of new and old classes (\textit{Avg}), weighting them equally.
We can see that fine-tuning shows an impressive forgetting on old classes, both on the 19-1 and 15-5 settings. Introducing the unbiased classification loss (\cref{eq:uce} helps in alleviating forgetting but the results are still low on old classes, clearly indicating that introducing a technique to prevent forgetting is required.
ILOD and FasterILOD, in fact, improve the performances on old classes. However, forgetting is prevented at the cost of a decrease in performance on novel classes: they both loses nearly 8\% on the 19-1 and 5\% on the 15-5 with respect to fine-tuning.
Differently, employing our proposed MMA we clearly improve the performance, preventing forgetting while showing good performance on novel classes. In particular, \wrt ILOD and Faster ILOD, MMA obtains, on new classes, nearly +5\% and +3\%, respectively on 19-1 and 15-5, while showing comparable performance on old classes.
Considering the extended version of MMA (MMA + $\ell_{dist}^{MASK}$), it slightly improves the performance on old classes \wrt MMA, while obtaning comparable results on the new ones. Overall, it obtains 41.1\% and 38.2\% on the 19-1 and 15-5, respectively, 0.3\% and 0.9\% better than MMA. Interestingly, we note that, without any regularization on the mask head (MMA), we can still achieve good segmentation performance. This is due to the non competitiveness among classes on the mask head, which only regress a binary segmentation mask, while the class is predicted by the classification head, as in standard Faster R-CNN.
Overall, MMA and its extension demonstrate to outperform the other baselines in instance segmentation, showing a good trade-off between learning the new classes and avoiding to forget the old ones.

\begin{table}[t]
\caption{Ablation study of the contribution of MMA components in the 15-5 setting. Results are mAP@0.5\%. MMA is in green.}
\setlength{\tabcolsep}{5pt}
\centering
% \begin{adjustbox}{width=\linewidth}
\begin{tabular}{ccc|ccccc}
\toprule
%\textbf{Method}              &
\textbf{\cref{eq:uce}} & \textbf{$\ell_{dist}^{RCN}$} & \textbf{$\ell_{dist}^{RPN}$} & \textbf{1-15} & \textbf{16-20} & \textbf{1-20} & \textbf{Avg} \\ \hline

-                 &  -      & -       & 14.2         & 59.2         & 25.4         & 36.7         \\
\cmark            &  -      & -       & 40.0         & 57.8         & 44.4         & 48.9         \\
% -               & l2      & -       & 72.5         & 58.0         & 68.9         & 65.3         \\
%  -              & l2      & \cmark  & 73.5         & 55.6         & 69.1         & 64.6         \\
\cmark            & UKD     & -       & 67.3         & 60.3         & 65.6         & 63.8         \\

\cmark            & l2      & \cmark  & \bf{73.7}    & 56.8         & 69.5         & 65.3         \\
\cmark            & CE      & \cmark  & 72.8         & 59.4         & 69.5         & 66.1         \\
\rowcolor{YellowGreen} 
\cmark            & UKD     & \cmark  & 73.0         & \bf{60.5}    & \bf{69.9}    & \bf{66.7}     \\ 
% \cmark          & UKD     & \cmark  & 68.3         & 60.1         & 66.2         & 64.2 \\
\toprule
\end{tabular}
% \end{adjustbox}
\label{tab:ablation}
\end{table}

\subsection{Ablation Study}
In Table \ref{tab:ablation} we report a detailed analysis of our contributions, considering 15-5 setting in incremental object detection. We ablate each proposed component: the unbiased classification loss (\cref{eq:uce}), the classification head knowledge distillation loss ($\ell_{dist}^{RCN}$), the use of the RPN distillation loss ($\ell_{dist}^{RPN}$), and finally, the use of a feature distillation loss, as proposed in \cite{peng2020faster}.
The first row indicates fine-tuning the network on the new data, without applying any regularization. It can be noted that the performances are poor on the old classes, while it achieves good performance on the new ones.
Adding the unbiased classification, the performance on the old classes substantially improves: from 14.2\% to 40.0\%. This is due to the handling of missing annotation that alleviates forgetting.
Introducing the unbiased distillation loss in \cref{eq:ukd} (UKD), the performances improves significantly, both on old classes, reaching 67.3\%, and new classes, going from 57.8\% to 60.3\%. We argue that the performances on the new classes improves thanks to the distillation loss since the model learns to better distinguish the old classes from the new ones, improving the overall precision, We then introduce the RPN distillation loss, obtaining the final MMA model. We see that the performance further improves on old classes, achieving 73.0\%, while the performance on the new classes is comparable.

Finally, we compare the unbiased knowledge distillation in MMA with other possible choices. Inspired by previous works we employ the L2 loss on the normalized classification scores \cite{shmelkov2017incremental, peng2020faster} and the cross-entropy (CE) loss between the probability of old classes \cite{li2017learning}. We see that MMA distillation outperforms them, especially on the new classes, clearly demonstrating that modeling the missing annotations is essential to properly learn them. Overall, MMA achieves on the average of old and new class performance 66.7\%, 1.4\% and 0.6\% more than using the L2 loss or the cross-entropy loss.
%Finally, we add to MMA the feature distillation loss proposed in \cite{peng2020faster}. Interestingly, we note that the performances suffers a drop on the old classes: from 73\% to 68.3\%. For this reason, we decided to not use the feature distillation.

% We start from three baselines: FT, Faster-ILOD \cite{peng2020faster} and ILOD \cite{shmelkov2017incremental}. We first add to FT our modified cross-entropy (UCE): this increases the ability to preserve old knowledge in all settings without harming too much performances on the new classes. With the addition of the latter performances grow by almost 26\% on old classes. Secondly, starting from \cite{peng2020faster} we add UCE and UKD, comparing the latter with a standard cross-entropy loss and an L2 loss as RoI knowledge distillation. UKD outperforms on the new classes the loss used by \cite{peng2020faster} by 3\%, demonstrating the effectiveness of such distillation loss. Moreover with this table we demonstrate the usefulness of the distillation loss on RPN and also that using a feature distillation as in \cite{peng2020faster} harms the model, achieving similar results on new classes while lowering the MAP on old classes by more than 3\%.

% \begin{figure*}[H]
% \centering
% \subfloat{
%   \includegraphics[width=87mm]{images/10-1.png}
% }
% \subfloat{
%   \includegraphics[width=87mm]{images/10-2.png}
% }
% \hspace{0mm}
% \subfloat{
%   \includegraphics[width=87mm]{images/10-5.png}
% }
% \subfloat{
%   \includegraphics[width=87mm]{images/15-1.png}
% }
% \caption{caption}
% \end{figure*}

\section{Conclusions}
\label{sec:conclusion}
We studied the incremental learning problem in object detection considering an issue mostly overlooked by previous works. 
In particular, in each training step only the annotation for the classes to learn is provided, while the other objects are not considered, leading to many missing annotations that mislead the model to predict background on them, exacerbating catastrophic forgetting. 
We address the missing annotations by revisiting the standard knowledge distillation framework to consider non annotated regions as possibly containing past objects. 
We show that our approach outperforms all the previous works without using any data from previous training steps on the Pascal-VOC 2007 dataset, considering muliple class-incremental settings.
Finally, we provide a simple extension of our method in the instance segmentation task, showing that it outperforms all the baselines.
We hope that our work will set a new knowledge distillation formulation for incremental object detection methods. We leave extending our formulation to one-stage detectors as a future work.

%%%%%%%%% REFERENCES
{\small
\bibliographystyle{ieee_fullname}
\bibliography{bib}
}

\end{document}